\documentclass[runningheads]{llncs}

\usepackage{eccv}

\usepackage{eccvabbrv}
\usepackage{authblk}
\usepackage{graphicx}
\usepackage{booktabs}
\usepackage{footnote}
\usepackage[accsupp]{axessibility}  

\usepackage[pagebackref,breaklinks,colorlinks,citecolor=eccvblue]{hyperref}
\usepackage{pifont}
\usepackage{booktabs}
\usepackage{arydshln}
\usepackage{color}
\usepackage{colortbl}
\usepackage{float}
\usepackage{rotfloat}
\usepackage{longtable}
\usepackage{diagbox}
\usepackage{enumitem}

\newcommand{\ourmodel}{BPO\xspace}

\usepackage{amsmath,amsfonts,bm}

\def\eqref#1{Equation~(\ref{#1})}

\def\1{\bm{1}}

\RequirePackage{amsmath}
\RequirePackage{amssymb}
\RequirePackage{amsthm}
\RequirePackage{bm} 
\RequirePackage{url}
\usepackage{multirow}
\usepackage{graphicx}
\usepackage{makecell}
\usepackage{booktabs}
\usepackage{array}
\usepackage{url}
\usepackage{algorithm}
\usepackage{algorithmic}
\usepackage{dsfont}

\usepackage{enumitem}

\usepackage{enumerate}

\usepackage{mathrsfs}

\def\$#1\${\begin{align*}#1\end{align*}}

\newcommand{\cD}{\mathcal{D}}

\newcommand{\EE}{\mathbb{E}}

\newcommand{\PP}{\mathbb{P}}

\usepackage{xcolor}

\definecolor{red1}{HTML}{f47983}
\definecolor{blue1}{HTML}{3eede7}
\definecolor{yellow1}{HTML}{f5dd6f}

\newcommand{\KL}{D_{\mathrm{KL}}}

\usepackage[skins]{tcolorbox} 
\tcbuselibrary{breakable} 
\newtcolorbox[auto counter, number within=section, list type=subsubsection, list inside=toc]{sectionbox}[2][]{
colback=white!98!gray, colframe=black, 
colbacktitle=white!90!gray, coltitle=black, 
fonttitle=\bfseries,
title={#2}, 
list entry={Comment \thetcbcounter\quad}
}
\usepackage{orcidlink}

\title{
Strengthening Multi-modal Large Language Model with Bootstrapped Preference Optimization} 

\title{
\raisebox{-0.25cm}{\includegraphics[width=1.0cm]{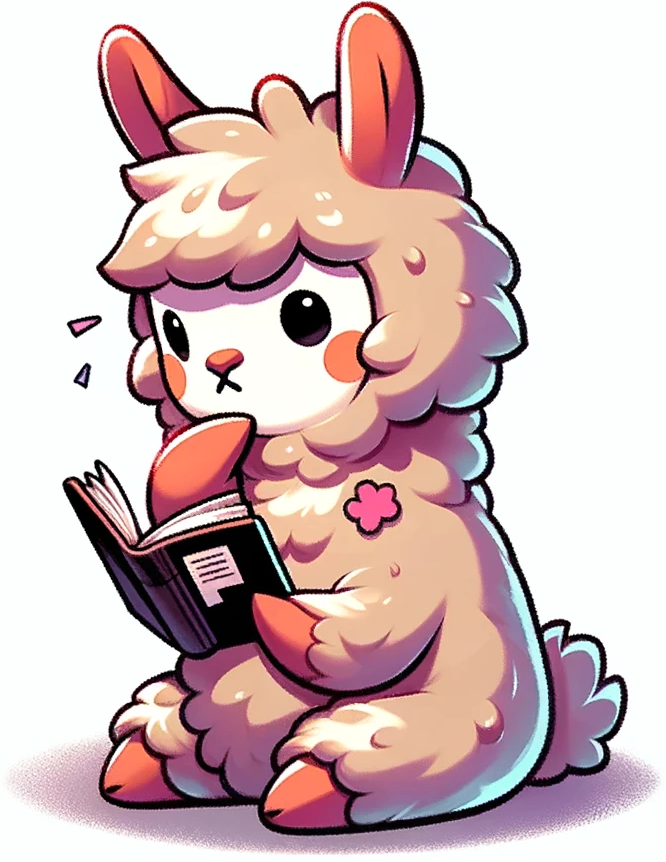}}
Strengthening Multimodal Large Language Model with Bootstrapped Preference Optimization} 

\titlerunning{Bootstrapped Preference Learning}

\author{Renjie PI\inst{1}\thanks{\, Equal Contribution. 
} \and
Tianyang Han\inst{3}$^*$ \and
Wei Xiong\inst{2} \\ 
Jipeng  Zhang\inst{1}\and Runtao Liu\inst{1} \and Rui Pan\inst{1} \and Tong Zhang\inst{2}}

\authorrunning{PI et al.}

\institute{The Hong Kong University of Science and Technology \and
University of Illinois at Urbana-Champaign \and
The Hong Kong Polytechnic University
}
\begin{document}
\maketitle

\begin{abstract}
  
Multimodal Large Language Models (MLLMs) excel in generating responses based on visual inputs. However, they often suffer from a bias towards generating responses similar to their pretraining corpus, overshadowing the importance of visual information. We treat this bias as a ``preference'' for pretraining statistics, which hinders the model's grounding in visual input. To mitigate this issue, we propose Bootstrapped Preference Optimization (BPO), which conducts preference learning with datasets containing negative responses bootstrapped from the model itself. Specifically, we propose the following two strategies: 1) using distorted image inputs to the MLLM for eliciting responses that contain signified pretraining bias; 2) leveraging text-based LLM to explicitly inject erroneous but common elements into the original response. Those undesirable responses are paired with original annotated responses from the datasets to construct the preference dataset, which is subsequently utilized to perform preference learning. Our approach effectively suppresses pretrained LLM bias, enabling enhanced grounding in visual inputs. Extensive experimentation demonstrates significant performance improvements across multiple benchmarks, advancing the state-of-the-art in multimodal conversational systems.
\end{abstract}

\section{Introduction}
The emergence of Large Language Models (LLMs) has marked a significant milestone in the field of AI, revolutionizing natural language processing and understanding~\cite{openlm2023openllama,openai2023gpt4, touvron2023llama, scao2022bloom, chowdhery2022palm, alpaca, vicuna2023}. These models, trained on vast text corpus datasets, possess rich world knowledge, making them excel in generating helpful and contextually relevant text. With the advancement of LLMs, Multimodal Large Language Models (MLLMs) have seen rapid improvements~\cite{liu2023llava, zhu2023minigpt4, su2023pandagpt, dai2023instructblip, li2023blip2, openai2023gpt4, bai2023qwenvl}, which typically process the images using a pretrained visual encoder (e.g., vision transformer) and feed them to the LLM as token embeddings along with the text token embeddings. These models extend the capabilities of LLMs to engage in interesting conversations with image inputs, which enables various potential applications such as autonomous driving~\cite{ding2023hilm} and medical assistants~\cite{li2023llavamed}.

\begin{figure*}[htp!]
\centering
\includegraphics[width=1\textwidth]{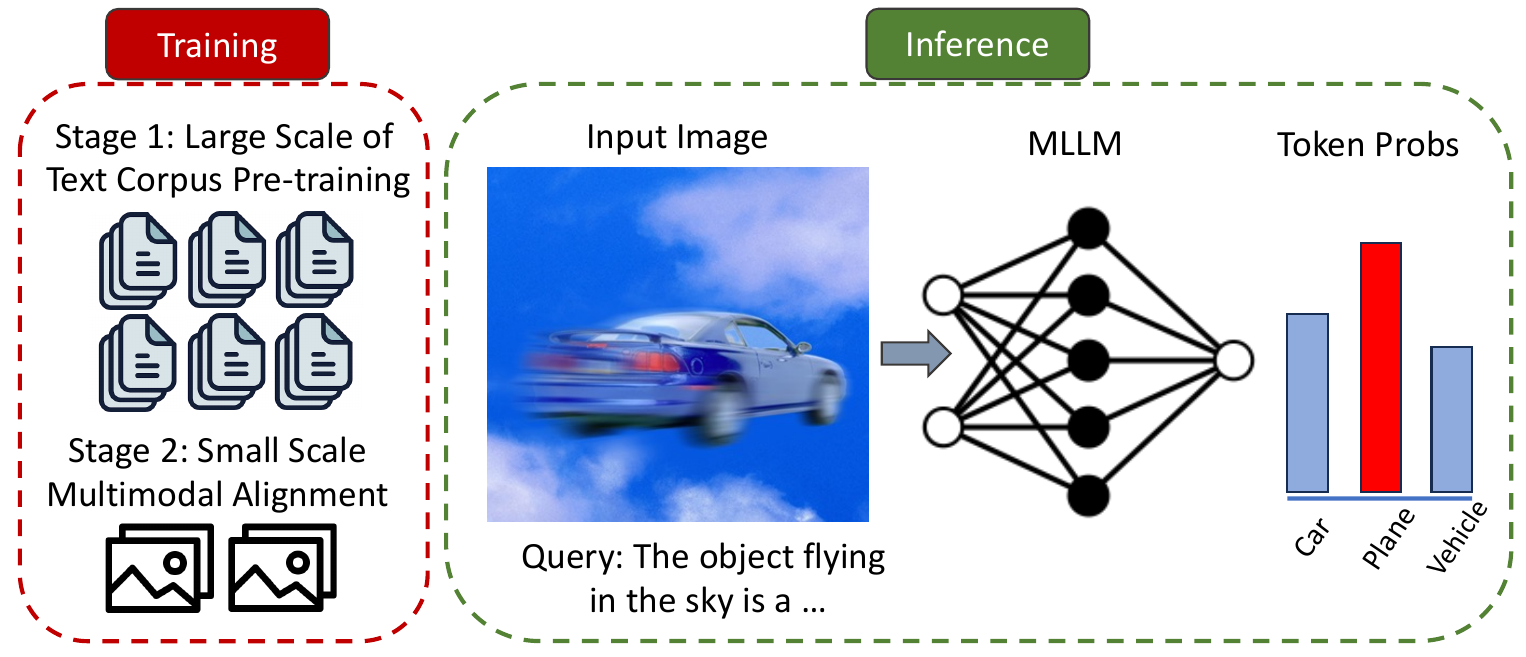} 
\vspace{-0.8cm}
\caption{Illustration of pretraining bias during MLLM's inference. Due to the difference in data scales between text-based pretraining and multimodal alignment, the MLLM is prone to generating contents that are frequently seen during its pretraining stage.}\label{fig:teaser}
\end{figure*}

Despite the fascinating capabilities of state-of-the-art Multimodal Large Language Models (MLLMs), they exhibit a susceptibility to producing erroneous or hallucinatory responses that do not correspond to the input image. For instance, MLLMs often generate non-existent objects, incorrectly identify attributes such as shape or color, or provide inaccurate object counts. This issue renders MLLMs unreliable and impractical for real-world applications, particularly those with high stakes, such as autonomous driving systems~\cite{ding2023hilmd} or medical assistants~\cite{li2023llavamed}.

We hypothesize one of the major causes of this phenomenon is the bias inherited from LLM's pre-training stage. Inspired by recent research in jailbreaking of LLMs~\cite{douglas2024mitigating}, we point out that MLLMs can be treated as mixture models, consisting of both distributions learned from the pretraining text corpus, as well as multi-modal alignment tuning. Specifically, the LLM undergoes an extensive pretraining stage with the large scale text corpus. Comparatively, the multi-modal alignment stage in current SOTA MLLMs utilizes much fewer training samples and shorter training period. The gap between the training scales of the two phases inevitably makes the pretraining distribution dominate the generation of MLLM under certain scenarios, especially when the image is of lower quality or is not sufficiently trained during multi-modal alignment.

Motivated by the reasons above, we introduce a novel stand point to tackle the aforementioned problem. Our study draws an analogy between a blind person who, even after a cornea transplant, still instinctively prefers walking on tactile paving. We argue that the distribution bias of MLLM stemming from pretraining can be viewed as an inherited ``preference'' derived from past prevalent behavior. Conversely, generating responses based on image inputs represents a new ``preference'' that the model must adapt to. To effectively address the current challenges faced by MLLMs, we propose to use the preference learning techniques from reinforcement learning (RL)~\cite{christiano2017deep, ziegler2019fine}, which is the leading technique to adapt the model generation toward the goals of being preferred. The effectiveness of the preference learning has been showcased with its tremendous success in Chat-GPT~\cite{openai2023gpt4}, Claude~\cite{bai2022training}, and Gemini~\cite{gemini2023}, and is known to be far more efficient than the SFT~\cite{ramamurthy2022reinforcement}. The primary goal of this paper is to extend these techniques to align the different modality of MLLMs. Specifically, the most standard and popular preference learning~\cite{ouyang2022training, bai2022training, touvron2023llama} consists of three steps:
\begin{itemize}
    \item construct a preference dataset, which consists of a pair of samples and the preference signal indicating which one is more preferred;
    \item model a reward function based on the preference dataset;
    \item optimize the reward function using proximal policy optimization (PPO) ~\cite{schulman2017proximal}.
\end{itemize}
While there are a diverse set of preference datasets in the LLMs, the preference learning in MLLMs is largely under-explored. To this end, our first contribution is an innovative strategy to obtain comparison pairs based on existing datasets with ground truth annotations. Specifically, we regard the existing datasets with ground truth annotations as positive responses, and generate negative responses by 1) Image-weakened prompting: we utilize distorted images as "weakened visual prompts" to elicit responses from the MLLM, revealing the inherent bias from pretraining. These responses contain a higher degree of erroneous patterns and align more closely with the pretraining distribution, while still being relevant to the image input. 2) LLM bias injection, we leverage the LLM component of the MLLM to directly modify the original responses using carefully designed prompts and few-shot examples, resulting in negative responses that exhibit similarities but differ in specific details from the original annotations. This collection of negative responses reveals a more pronounced bias towards the pretraining distribution, thereby exposing potential weaknesses and unreliability of the MLLM.

In terms of algorithmic design, it is known that the PPO algorithm is unstable and sample-inefficient in aligning LLMs~\cite{choshen2019weaknesses} and imposes a heavy burden on the GPU resources as it requires loading multiple (typically four) models at the same time~\cite{yuan2023rrhf, dong2023raft}. In contrast, the recently proposed direct preference optimization (DPO) combines the reward modeling with the policy optimization into one step, and directly learns from the preference dataset (hence the name). The DPO algorithm has emerged as a promising alternative to RLHF due to its stability and competitive performance. Motivated by this, we propose a variant of DPO, referred to as the \textbf{Bootstrapped Preference Optimization (\ourmodel)}, to extend the techniques to the MLLMs, which can significantly boosts the model performance as evaluated by multiple popular visual understanding benchmarks, while reducing object hallucinations by a large margin. To summarize, we make the following contributions in this paper:
\begin{itemize}
    \item Firstly, make take a novel view and formulate the multimodal alignment into preference learning task, where the pretraining bias and visual grounding are treated as the old and new preferences, respectively.
    \item Secondly, we introduce a novel approach to construct preference datasets automatically at scale. The collected negative samples effectively expose the pretraining bias of MLLM.
    \item Lastly, we demonstrate through empirical evidence that our approach effectively enhances the grounding of MLLM on image inputs and results in performance boost in multiple benchmarks.
\end{itemize}
\label{sec:intro}
\begin{figure*}[htp!]
\centering
\includegraphics[width=1\textwidth]{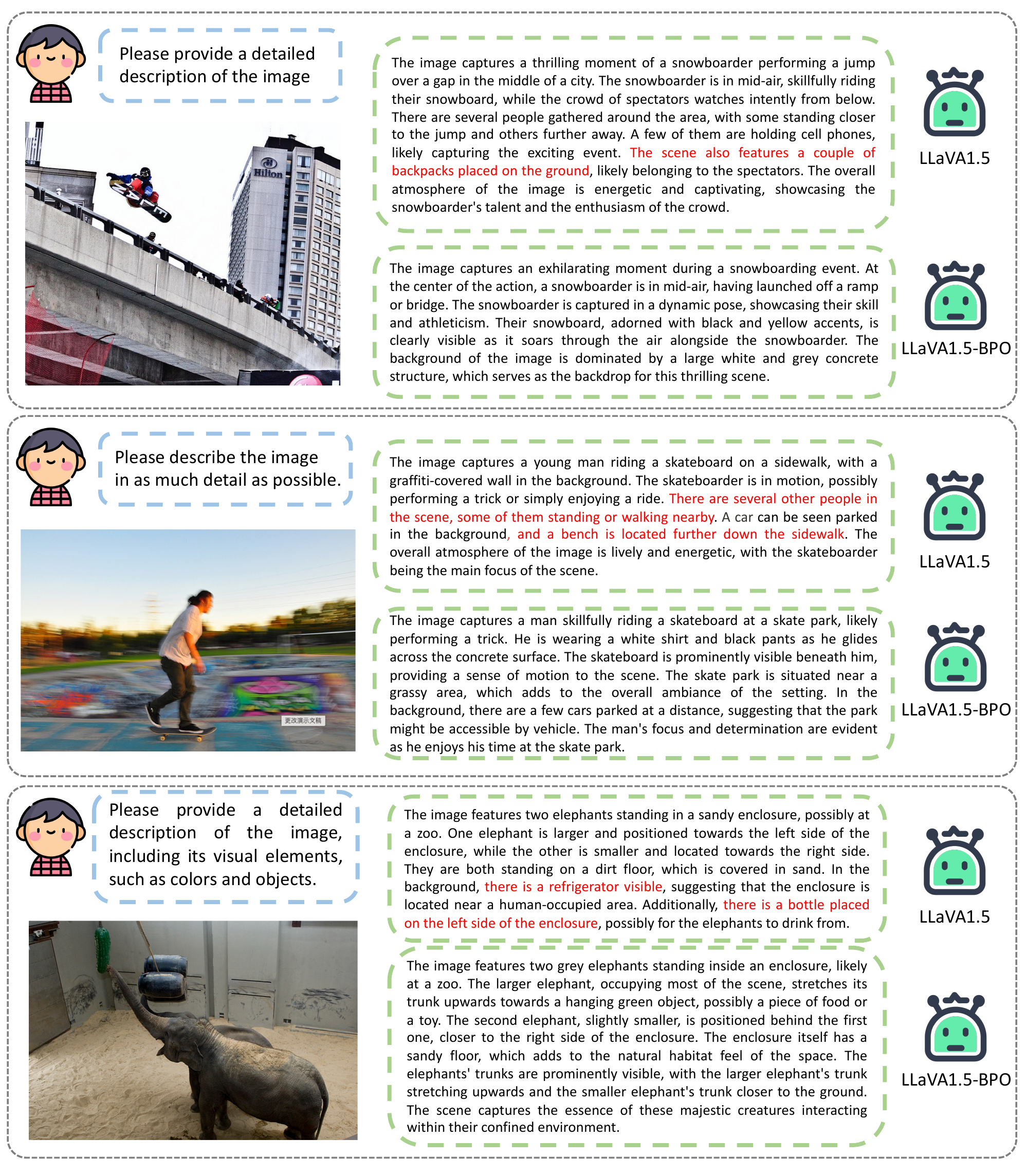} 
\vspace{-0.8cm}
\caption{We demonstrate a few examples of responses generated before and after \ourmodel. The responses generated by the MLLM after \ourmodel improves the grounding on visual inputs, which improves visual faithfulness and results in less erroneous outputs.}\label{fig:qualitative}
\end{figure*}
\section{Related Work}
\paragraph{Multi-Modal Large Language Model.}
In recent years, transformative advancements have been witnessed in the development of large language models (LLMs)~\cite{brown2020language, scao2022bloom, chowdhery2022palm, smith2022using, hoffmann2022training, ouyang2022training, touvron2023llama, bai2022training}. These advancements have greatly elevated the capabilities of language understanding and generation, showcasing near-human proficiency across diverse tasks. Concurrently, the success of LLMs has inspired explorations into the incorporation of visual modality into LLM, leading to the emergence of multi-modal large language models (MLLMs)~\cite{liu2023llava, li2023blip2, dai2023instructblip, zhu2023minigpt4, dai2023instructblip, openai2023gpt4, bai2023qwenvl, su2023pandagpt, gao2023llamaadapter, pi2023detgpt, pi2023perceptiongpt, gao2023gllava}. These models have demonstrated remarkable abilities in engaging in dialogue based on visual inputs.

\paragraph{Alignment of Large Language Model.}
Alignment in agent behavior, originally introduced by \cite{leike2018scalable} ensures that actions align with human intentions. Reinforcement Learning from Human Feedback (RLHF) approaches, such as those presented in \cite{ouyang2022training, stiennon2020learning, nakano2021webgpt, bai2022training, bai2022constitutional, glaese2022improving, ziegler2019fine, wu2021recursively, scheurer2023training}, utilize methods like PPO \cite{schulman2017proximal} to maximize the rewards of model outputs. The successful alignment of InstructGPT in GPT-3, as described in \cite{brown2020language}, also involves supervised fine-tuning (SFT). In the domain of visual models, alignment studies, such as those explored in \cite{hao2022optimizing, lee2023aligning, wu2023better}, focus on interpreting specific visual signals~\cite{lee2023aligning}, with ongoing challenges in balancing human preferences and image fidelity. RRHF~\cite{yuan2023rrhf} and RAFT~\cite{dong2023raft, diao2023lmflow,wang2024arithmetic} leverage the capabilities of large language models (LLMs) to bootstrap responses, followed by fine-tuning the model on the subset of collected samples with high rewards. \cite{rafailov2023direct} propose direct preference optimization (DPO), which directly learns from the offline dataset with a clever reparameterization technique. The DPO is later extended to the online setting by  Xiong. et. al~\cite{xiong2023gibbs}.  Recently, several works have investigated the vulnerability of MLLM against malicious image inputs~\cite{liu2023queryrelevant, pi2024mllmprotector} More recently, Silkie\cite{li2023silkie} suggests curating preference data to fine-tune multi-modal large language models (MLLMs) using responses generated by a pool of different MLLMs. These responses are paired with image and textual queries and subsequently scored by GPT4-V, which necessitates a large number of GPT4-V API calls and lacks scalability.

\begin{figure*}[htp!]
\centering
\includegraphics[width=1\textwidth]{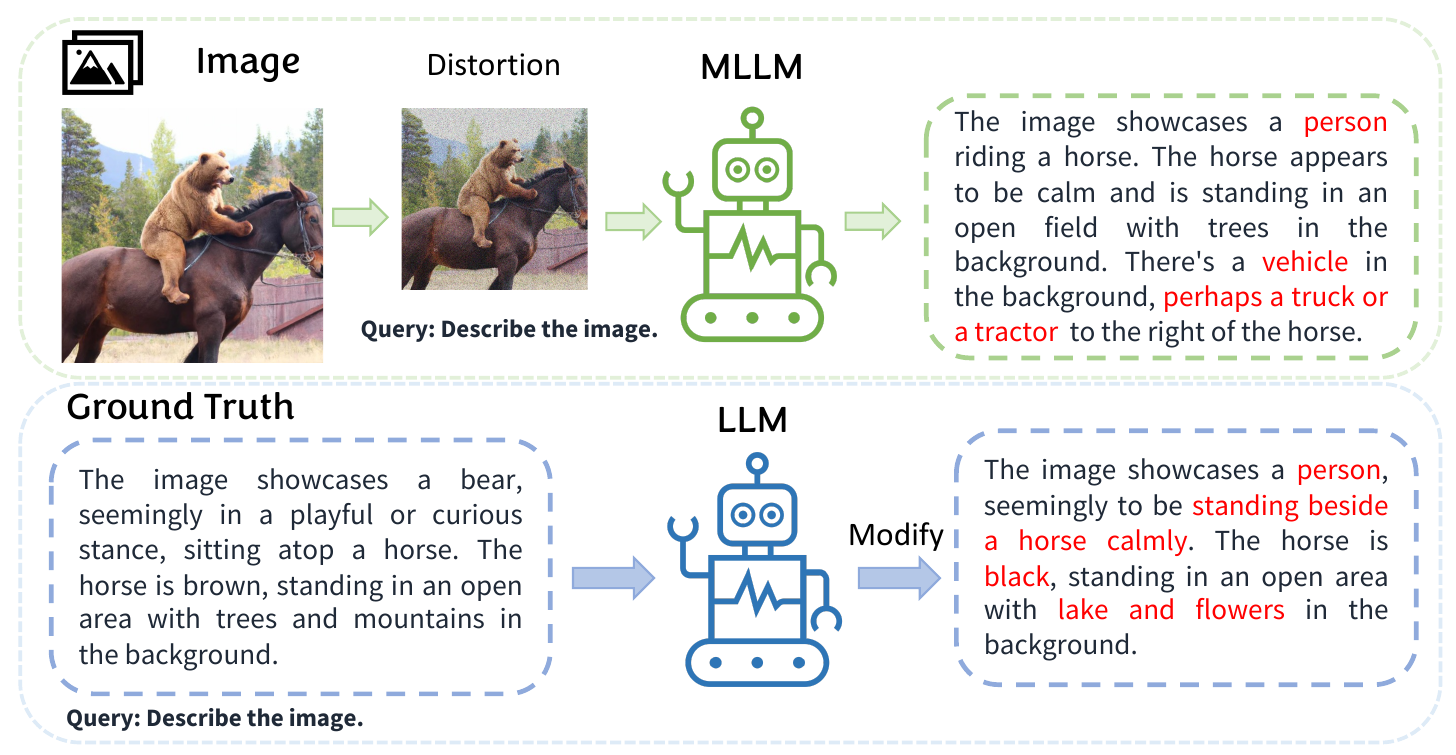} 
\caption{The generation pipeline for negative response. Top: Image weakened prompting, which elicits responses containing pretraining bias by injecting noises into the image features; Bottom: LLM-bias injection, which explicitly modifies the details of the ground truth responses by injecting erroneous but common elements.}\label{fig:framework}
\end{figure*}

\paragraph{Hallucination in Multimodal Large Language Models.}
Recently, many efforts have been dedicated to alleviate the hallucination of MLLMs, which may manifest across various aspects, including object existence, object count, attribute, and relation between objects~\cite{li2023evaluating, liu2023mitigating, han2024instinctive}. Woodpecker~\cite{yin2023woodpecker} leverages external tools such as object detectors and LLM to correct the hallucination in MLLM's responses. Despite the effectiveness, the external tools make this method inflexible and does not improve the MLLM's true performance. LRV-Instruction~\cite{liu2023mitigating} proposes to conduct supervised fine-tuning (SFT) on the MLLM with positive and negative instructions that focus on the semantics of objects. However, SFT on these instructions hinder the capability of MLLMs for generating detailed responses. Visual contrastive decoding~\cite{leng2023mitigating} proposes to correct the MLLM's output bias by subtracting the output from the MLLM using distorted image as input, which requires inference two models for each token. LLaVA-RLHF~\cite{sun2023aligning} introduce RLHF into MLLM training pipeline by training a reward model, which reduces hallucinatin via PPO. However, this approach requires manually labelled data for reward model training.

\paragraph{Generating Training Data from LLM.}
Owing to the advent of powerful large language models (LLMs), a new line of research that aim to automatically bootstrapping training data from LLMs has drawn lots  of attention. For instance, several works propose generating training data from a more powerful LLM to tune another student language model~\cite{meng2022generating, meng2023tuning,ye2022zerogen}. Other works propose to synthesize data with better quality using more advanced techniques~\cite{gao2023selfguided, meng2023tuning}, such as bi-level optimization. Recently, LLMs are used to synthesize data to improve the model's reasoning ability~\cite{yu2023metamath,  gao2023gllava}. In our work, we propose leveraging the model to bootstrap negative responses for preference learning.
\label{sec:related_work}

\begin{figure*}[htp!]
\centering
\includegraphics[width=1\textwidth]{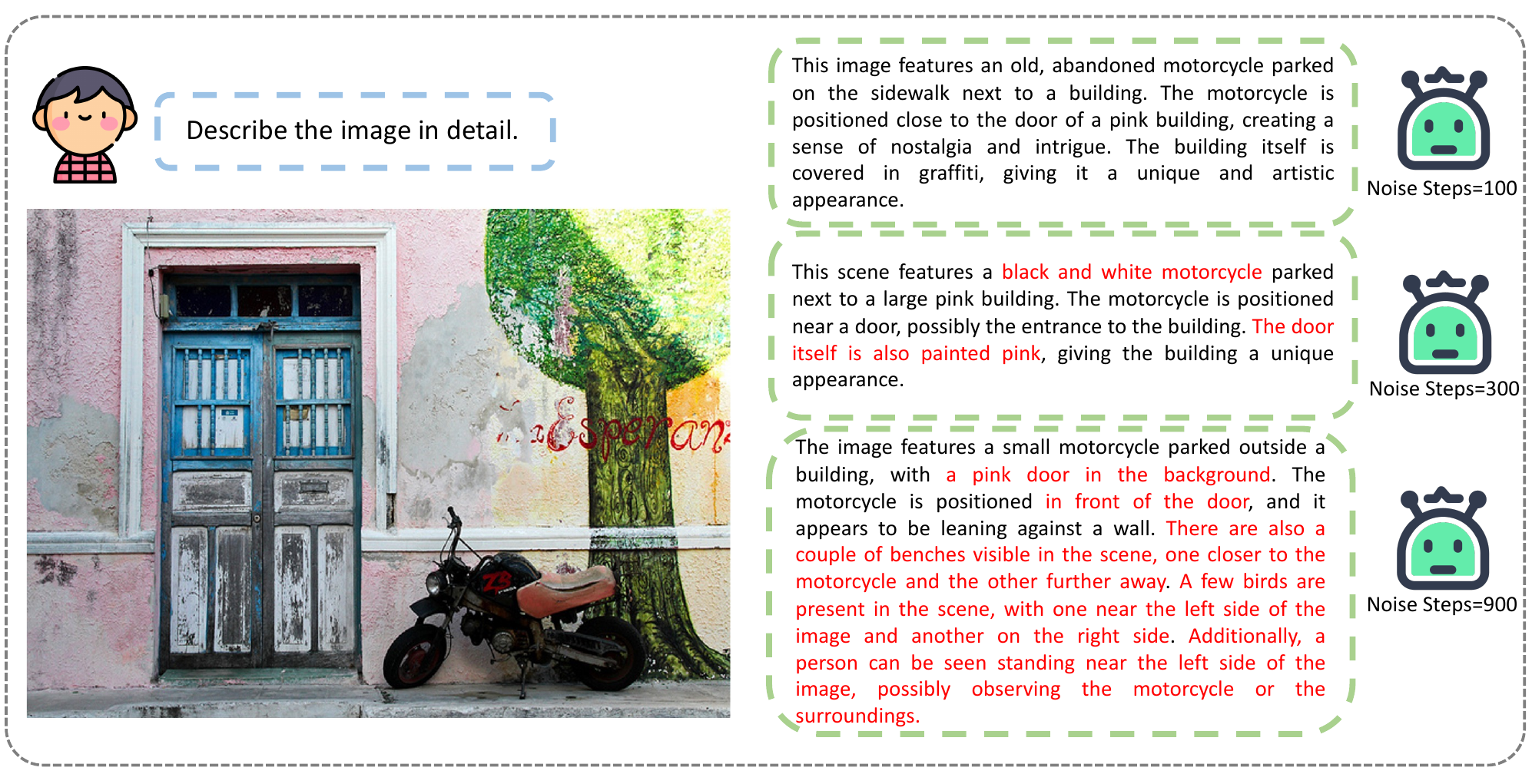} 
\vspace{-0.8cm}
\caption{The MLLM-generated responses with continuously growing steps of added noise. We can see that higher level of noise leads to more decline in visual faithfulness  and the generation of hallucinated objects. These responses expose the over-reliance on knowledge learning from pretraining corpus, which is leveraged to suppress the unwantede pretraining bias via our \ourmodel.}\label{fig:image-weakened-responses}
\end{figure*}

\section{Scalable Preference Dataset Generation}
A preference dataset $\cD$ consists of numerous tuples, such as $(I, q, r^1, r^2, p)$, where  $I$ is the image, $q$ is the query, $r^1$ and $r^2$ are the two responses, and $p$ is the preference signal where $p=1$ indicates that $r^1 \succ r^2$ given $(I,q)$, while $p=0$ stands for $r^1 \prec r^2$.

Annotating preference datasets manually can be a laborious and time consuming process. For instance, previous work~\cite{sun2023aligning, yu2023rlhfv} hire crowd workers to identify potential hallucinations in the model’s responses, where the responses associated with less hallucination are assigned with higher scores. The labelled responses are subsequently leveraged to construct the preference dataset for training the reward model. This costly human-labelling process prohibits the scalability of such approaches. 

On the other hand, there are abundant existing datasets targeted for supervised fine-tuning, which are annotated with high-quality image-question-answer triplets. For example, LLaVA~\cite{liu2023llava} and MiniGPT4~\cite{zhu2023minigpt4} utilize the fine-grained annotations (e.g., captions, bounding boxes) to generate high-quality captions and QA pairs that are associated with images. ShareGPTV~\cite{chen2023sharegpt4v} leverages the powerful GPT4-V to produce high-quality captions for images. Given that the annotations in these high-quality datasets are well grounded to the image contents, they can readily serve as the positive responses in preference pairs.

\begin{figure*}[htp!]
\centering
\includegraphics[width=0.8\textwidth]{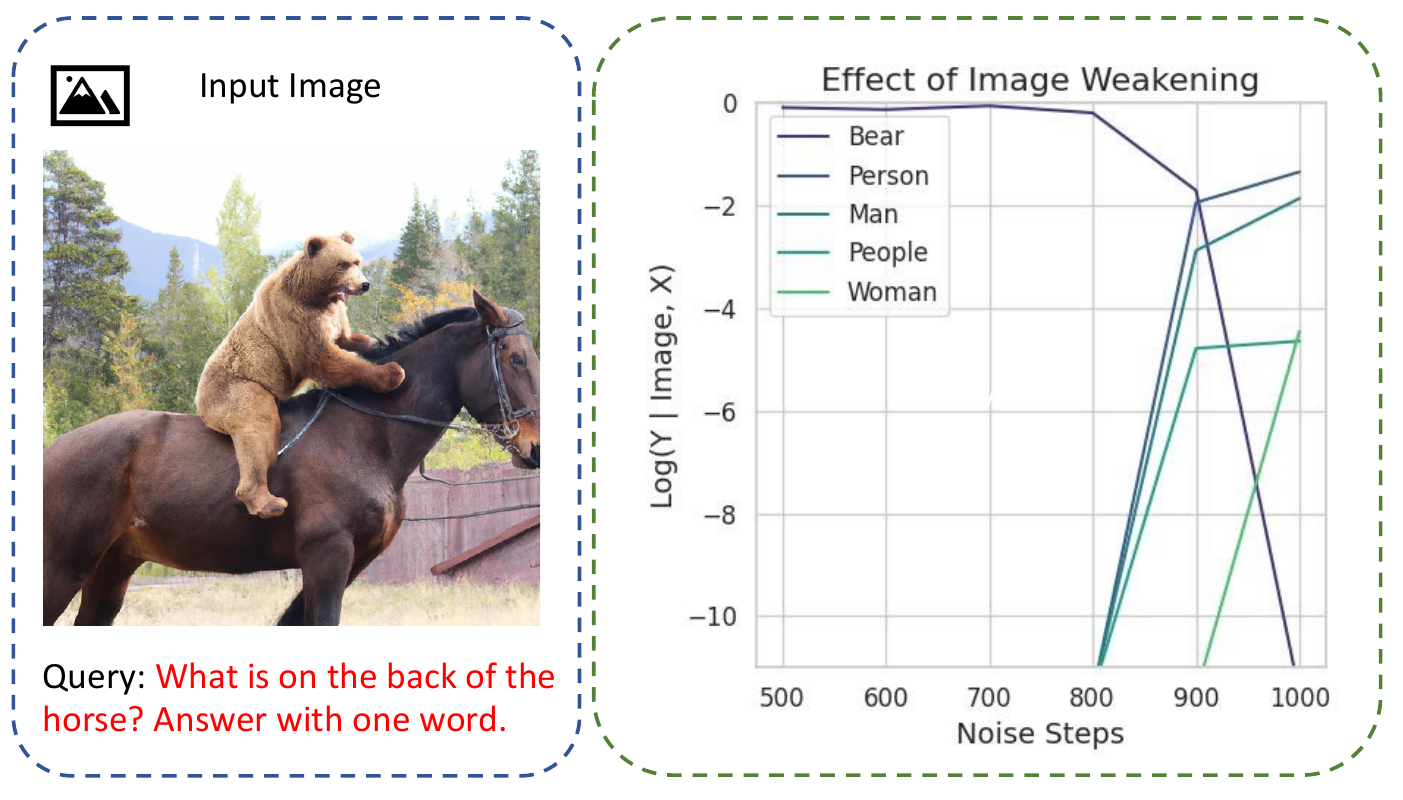} 
\vspace{-0.2cm}
\caption{The effect of image-weakened prompting. We observe the change in logits from the MLLM's output by continuously injecting higher level of noise into the image features. The log likelihood of the bear starts to decrease when the noise gets higher, and the likelihood of words such as ``Person'', ``Man'', ``People'' and ``Woman'' starts to increase, and finally take over ``bear''. This demonstrates the pretraining bias starts to overwhelm image information when the image is weakened.}\label{fig:image-weakened-prompting}
\vspace{-0.9cm}
\end{figure*}

\subsection{Negative Response Collection}
 To automatically collect negative responses at scale without excessive cost, we propose the following strategies.

\textbf{Image-Weakened prompting:} To expose the pretraining bias and potential weaknesses of MLLMs, we apply distortions to the image features before providing them to the MLLMs for inference. Specifically, inspired by ~\cite{leng2023mitigating}, we apply gaussian mask on the image embeddings from CLIP model, which is analogous to the forward process of diffusion models~\cite{ho2020denoising}.

In the context of MLLMs, image input can be treated as a part of the prompt, which brings the MLLM's output distribution to the visual domain. After being applied with distortion, the strength of the image becomes weaker, which makes the model more likely to be overwhelmed by the pretraining distribution, further leading to inaccurate responses. As shown in figure~\ref{fig:image-weakened-prompting}, the MLLM becomes more likely to generate tokens commonly seen in pretraining corpus when the image is weakened by noise. Therefore, those responses can well expose the pre-training bias of the MLLM. We also visualize the responses generated using distorted image inputs in figure~\ref{fig:image-weakened-responses}.

\textbf{Error Injection:} We take a more direct approach by leveraging the LLM component of the MLLM to explicitly modify the ground truth responses. Specifically, we prompt the LLM to tweak the details in the responses, such that the modified responses are similar to the original ones, but different from some aspects (e.g., object existence, object attributes, object counts). We prompt the LLM to ensure the modified response is logical and common in reality, which is likely to be close to the pretraining distribution (prompts shown in table~\ref{prompt:error_injection}).

\begin{table*}[htp!]
\centering
\caption{Prompt for error injection via LLM component. We explicitly let the LLM component (Vicuna) of the MLLM (LLaVA) produce responses modified to include errors that are commonly seen.} 
\begin{minipage}{1.0\textwidth}\vspace{0mm}    \centering
\begin{sectionbox}[]{Prompt for Error Injection via LLM Component}
    \centering
      \footnotesize
    \begin{tabular}{p{0.97\textwidth} c}
I will give you a question and a response about an image. Pretend that you can see the image. You must modify the response by changing the original details, such as adding more objects or change the attributes of objects. Note that the modified responses should still be common in reality. I will give you some examples, then you need to paraphrase the new response.\\

\textbf{Examples}:\\

Question: Describe the image. \\
Response: the image shows an old man walking across the street. There are many people on the sidewalks, and cars are running on the street.  
\\Modified response: the image shows a woman running towards a car, there are no cars on the street.\\
\\
Question: What is the color of the apple?\\ Response: the color of the apple is black.\\
Modified response: the color of the apple is yellow.\\
\\
Question: What is on the left side of the boy? \\
Response: there is a dog left to the boy.\\
Modified response: it is a cat on the left of the boy.\\
\\
Question: How many people are there? \\Response: there are 3 people in the image. \\
Modified response: there are only 1 person in the image.\\
\\
Question: what is the price of the knife?\\ Response: The price shown in the image is 3 dollors. \\
Modified response: \\
\\
Question: Describe the image in detail. \\Response: The image is a close-up of a wristwatch with a circular face and a black leather strap. The watch has a silver-colored bezel and case, and features a white dial with black Arabic numerals indicating the hours from 1 to 12. The hands of the watch are not visible....\\
Modified response: The image is a close-up of a wristwatch with a rectangular face and a brown leather strap. The watch has a gold-colored bezel and case, and features a black dial with white Roman numerals indicating the hours from 1 to 12. The hands of the watch are pointing towards the 3 o'clock position...\\
\\
Question: \textcolor{red}{New Question}\\
Response: \textcolor{red}{ New Response}\\
Modified Response:
\end{tabular}
\end{sectionbox}
\label{prompt:error_injection}
\end{minipage}
\end{table*}

\subsection{Data Sources}
We collect ground truth annotations from the following datasets: 1) ShareGPT-V~\cite{chen2023sharegpt4v}: a captioning dataset constructed by prompting responses from the powerful GPT4-V, which contains detailed responses with rich visual concepts; 2) LLaVAR~\cite{zhang2024llavar}: a VQA dataset consisting of images with rich texts; 3) LLaVA-Instruct~\cite{liu2023llava}: the original instruction tuning dataset from LLaVA, which comprises of image-based conversations.
\begin{table*}[t]
\centering
\caption{Data sources of our preference dataset. We uniformly sample data from the popular visual instruction tuning datasets.}
\resizebox{1\textwidth}{!}
{
\begin{tabular}{llc}
\toprule
\textbf{Source}  & \textbf{Content} & \textbf{Samples Num} \\
ShareGPT-V~\cite{chen2023sharegpt4v} & High-quality image captioning dataset annotated by GPT4-V & 57906 \\
LLaVAR~\cite{zhang2024llavar} & VQA dataset consisting of images with rich texts & 55445 \\
LLaVA-Instruct~\cite{liu2023llava} & Instruction dataset comprising of image-based conversations & 54359 \\

\bottomrule
\end{tabular}}
\end{table*}

\section{Direct Preference Optimization}
To facilitate the preference learning, one common assumption is that Bradley-Terry (BT) model~\cite{bradley1952rank, ouyang2022training, rafailov2023direct, xiong2023gibbs}, which states that there exists a reward function $\phi^*(I,q,r) \to [0,1]$ so that the preference satisfies 
\begin{equation} \label{eqn:bt} \small
\begin{aligned}
        \PP(r^1 \succ r^2|x,r^1,r^2)
        &=  \frac{\exp(\phi^*(I,q,r^1))}{\exp(\phi^*(I,q,r^1)) + \exp(\phi^*(I,q,r^2))}\\
 &= \sigma\big(\phi^*(I,q,r^1)-\phi^*(I,q,r^2)\big),
 \end{aligned}
\end{equation}
where $\sigma(z) = 1/(1+\exp(-z))$ is the sigmoid function. Essentially, the BT model implies that the preference probability is a non-decreasing and non-linear transformation (the sigmoid function) of the reward difference. This also partially explains why we choose to generate negative samples by error injection or imag-weakened prompting. Otherwise, if the two samples are of similar quality, even the preference signal from the human can also be noisy, which may hurt the subsequent preference learning. 

Under the BT model, the learning objective of preference learning~\cite{ouyang2022training, bai2022training, rafailov2023direct, xiong2023gibbs} is 
\begin{equation}
\label{eqn:value}
    J(\pi)
    = \EE_{I,q \sim d_0} \Big[\underbrace{\EE_{r \sim \pi(\cdot|I,q)}[\phi^*(I, q, r)]}_{\displaystyle \text{Optimize the reward}} - \underbrace{\eta \KL(\pi_\theta(\cdot|I,q)\Vert \pi_0(\cdot|I,q))}_{\displaystyle \text{Stay close to the initial model}} \Big],
\end{equation}
where $\eta>0$ is the KL penalty coefficient, $\pi_0$ is the initial model,  $\pi_\theta$ is the model to optimize. After obtaining the preference dataset $\cD = \{(I, q,r^1,r^2,p)\}$ , in the classic framework~\cite{ouyang2022training, bai2022training, touvron2023llama},  two stages of training can be performed: 1) the reward model $\pi_\theta$ can be trained under the Bradley-Terry (BT) model; 2) train the model $\pi_\theta$ with online RL algorithm, such as proximal policy optimization (PPO)~\cite{schulman2017proximal}. However, previous research has found that training the reward model with multi-modal inputs leads to more severe reward hacking~\cite{sun2023aligning}, since the continuous nature of image inputs makes the modelling of preference more challenging. Meanwhile, the instability of DRL-based PPO requires extensive efforts to tune the model to its best performance. 

Recently, a more easy-to-tune approach has been proposed for aligning the preference, which is termed  Direct preference optimization (DPO). This method prevents the need for training an external reward model by directly fine-tuning it on an offline preference dataset. The key insight is that the maximization of \eqref{eqn:value} admits a computationally intractable solution~\cite{zhang_2023_ltbook}:
\begin{equation*}
    \pi^*(r|I,q) = \frac{1}{Z(I,q)} \pi_0(r|I,q) \exp(\frac{1}{\eta} \phi^*(I,q,r)),
\end{equation*}
where $Z(I,q) = \sum_{r'} \pi_0(r'|I,q) \exp(\frac{1}{\eta} \phi^*(I,q,r'))$ is the normalization constant that cannot be computed in practice. Then, we may solve the reward as 
\begin{equation} \label{eqn:repara}
    \phi^*(I,q,r) = \eta \log \frac{\pi^*(r|I,q)}{\pi_0(r|I,q)} + \eta \log Z(I,q).
\end{equation}
Plugging \eqref{eqn:repara} back into the Bradley-Terry model in \eqref{eqn:bt}, we can now conduct maximal likelihood estimation (MLE) in the policy space, i.e., the space of generative model, directly by minimizing the following loss function (the negative log-likelihood):
\begin{equation} \label{eqn:dpo_loss}
\begin{aligned}
 \sum_{(I, q,r_w,r_l) \in \cD} -\Big[ \log \sigma\Big(\eta \log \frac{\pi_{\theta}(r_w|I,q)}{\pi_0(r_w|I,q)} - \eta \log \frac{\pi_{\theta}(r_l|I,q)}{\pi_0(r_l|I,q)} \Big)\Big],
    \end{aligned}
\end{equation}
where $r_w$, $r_l$ is the positive (winning) and negative (losing) response, respectively. 

\noindent \textbf{Interpretation}  As shown in \eqref{eqn:dpo_loss}, the first term boosts the reference-normalized log-likelihood of the positive response, while the second term penalizes that of the negative response. Optimizing \eqref{eqn:dpo_loss} increases the margin between the positive sample and negative sample, thus improving the preference for grounding on visual inputs. One notable feature is the presence of the KL divergence, which is critical for preventing the model from overfitting and distribution collapse. Without the KL-penalty, the optimal policy of \eqref{eqn:value} is greedy and deterministic in terms of the reward function, which deviates from the principle of generative models and can lead to an inferior performance without additional regularization \cite{dong2023raft, lin2023speciality}. 


The DPO formulation is related to contrastive learning~\cite{chen2020simple} to some degree, which also leverages pairs of positive and negative samples, i.e., a sample's representation should be closer to its positive references, and further from negative references. The difference mainly lies in the following: contrastive learning is an unsupervised learning framework, which utilizes the samples' representation distances between their positive and negative references to optimize the decision boundary. On the other hand, DPO utilizes preference datasets with labelled preference dataset and directly optimize the model's output probability.
\label{sec:method}
\begin{table*}[htp!]
    \centering
 \vspace{-0.25cm}
 \caption{Results on MM-Vet and LLaVA-Wild benchmarks. We observe consistent performance boosts over baseline models across all tasks on the two benchmarks. Notably, our tuned LLaVA1.5-7B-BPO even surpasses the larger LLaVA1.5-13B baseline on majority of the tasks. }
    \resizebox{0.98\textwidth}{!}{%

\begin{tabular}{c!{\vrule width 0.5pt}cccccccc!{\vrule width 0.5pt}cc!{\vrule width 0.5pt}c}

\toprule
 & \multicolumn{8}{c}{MM-Vet} &\multicolumn{2}{c}{Object-HalBench}& LLaVA-Wild\\
Model & Rec & OCR & Know & Gen & Spat & Math & Total& & $\text{Resp}^{\downarrow}$& $\text{Obj}^{\downarrow}$&All  \\
\midrule

MiniGPT-4-8B \cite{zhu2023minigpt4} & 27.4 & 15.0 & 12.8 & 13.9 & 20.3 & 7.70 & 22.1 & &- &- &- \\
BLIP-2-12B \cite{li2023blip2} &  27.5 & 11.1 & 11.8 & 7.00 & 16.2 & 5.80 & 22.4& &- &- &38.1 \\
 LLaVA-7B \cite{liu2023llava} & 28.0 & 17.1 & 16.3 & 18.9 & 21.2 & 11.5 & 23.8 & &- &- &-\\
 MiniGPT-4-14B \cite{zhu2023minigpt4} & 29.9 & 16.1 & 20.4 & 22.1 & 22.2 & 3.80 & 24.4& &- &- &- \\
Otter-9B \cite{li2023otter} & 27.3 & 17.8 & 14.2 & 13.8 & 24.4 & 3.80 & 24.7& &- &- &- \\
InstructBLIP-14B \cite{dai2023instructblip} & 30.8 & 16.0 & 9.80 & 9.00 & 21.1 & 10.5 & 25.6 & &- &- & 58.2 \\
LLaVA-13B \cite{liu2023llava} & 30.9 & 20.1 & 23.5 & 26.4 & 24.3 & 7.70 & 26.4 & &63.0 &29.5 &67.3 \\
LLaVA1.5-7B \cite{liu2023llava} & 37.0 & 22.9 &16.8  & 20.2 &25.7 &7.70 &31.7 & & 45.9 &23.7 &63.8 \\
LLaVA1.5-13B \cite{liu2023llava} & 41.1 & 29.1 & 23.0 & 24.2 & 35.6 & 7.70 & 36.8& &45.2 & 21.8 &71.2 \\
\midrule
\textbf{LLaVA1.5-7B-BPO} \cite{liu2023llava} & \textbf{41.3} & \textbf{29.5} & \textbf{24.8} & \textbf{27.8} & \textbf{34.8} & \textbf{11.5} & \textbf{36.8} & &\textbf{31.9}  &\textbf{15.1} &\textbf{71.6} \\
\textbf{LLaVA1.5-13B-BPO} \cite{liu2023llava} & \textbf{46.9} & \textbf{31.6} & \textbf{34.6} & \textbf{37.2} & \textbf{36.1} & \textbf{11.5} &\textbf{41.4}& &\textbf{27.3}   &\textbf{12.9} &\textbf{74.4} \\
\bottomrule
\end{tabular}
}
\label{tab:helpfulness}
\end{table*}
\section{Experiments}
\subsection{Implementation Details}
We finetune the MLLM from checkpoints of LLaVA1.5~\cite{liu2023llava}. We adopt parameter efficient training technique to save computational cost and alleviate catastrophic forgetting. Specifically, we use LoRA with rank set to 64. We use learning rate of $2e^{-6}$ and train the model for 2 epochs.  The model is trained on 8 A40 GPUs with 48G memory each, the batch size per GPU is set to 4. The training takes around 17 hours to complete for 7B model, ad 28 hours for 13B model.
\subsection{Evaluation Benchmarks and Metrics}
\noindent \textbf{Helpfulness Evaluation.} We use the following benchmarks for evaluation of MLLM's helpfulness: 1) \textit{LLaVA-Bench}~\cite{liu2023llava} is a real-world benchmark consisting of 60 tasks for testing LLaVA's visual instruction-following and question-answering abilities in natural environments; 2) \textit{MM-Vet} ~\cite{yu2023mmvet} evaluates multi-modal understanding by measuring six core visual-language capabilities across 128 tasks. It offers a comprehensive assessment that combines math, reasoning, and visual knowledge; 

\noindent \textbf{Visual Truthfulness  Evaluation.} For evaluation of visual truthfulness, we leverage \textit{Object HalBench}~\cite{rohrbach2019object}, which aims to assess the MLLM's hallucination in their generated image descriptions, we follow~\cite{yu2023rlhfv} to apply 8 diverse prompts for providing detailed descriptions of images. We evaluate hallucinations at both the response level (the percentage of responses that contain hallucinations) and the object level (the percentage of hallucinated object mentions compared to all object mentions). For all benchmarks, we leverage the powerful GPT4 as judge.

\subsubsection{Qualitative Results}
We showcase a few examples of MLLM-generated responses before and after \ourmodel tuning in Table~\ref{fig:qualitative}. We observe that after \ourmodel tuning, the MLLM is able to produce responses that are more grounded with the visual inputs and contain less erroneous elements.
\subsection{Results on Visual Helpfulness and Truthfulness}
We evaluate the effectiveness of our proposed \ourmodel on the popular MM-Vet and LLaVA-Wild benchmarks for helpfulness, and Object-Hallucination bench for visual truthfulness in table~\ref{tab:helpfulness}. Compared with the baseline models, we observe consistent performance boosts across all tasks on the three benchmarks. Surprisingly, our tuned LLaVA1.5-7B-BPO even surpasses the larger LLaVA1.5-13B baseline on majority of the tasks. Therefore, after strengthening the preference of MLLM over visual inputs, both the helpfulness and truthfulness of the MLLM can be greatly boosted.

\begin{table*}[htp!]
    \centering
\caption{Comparison with SFT baselines. The SFT datasets are constructed by extracting the positive responses from the preference dataset.
}
{
\begin{tabular}{l!{\vrule width 0.5pt}cccccccc}

\toprule
Model & Rec & OCR & Know & Gen & Spat & Math & Total\\
\midrule
LLaVA1.5-7B & 37.0 & 22.9 &16.8  & 20.2 &25.7 &7.70 &31.7 \\
LLaVA1.5-13B & 41.1 & 29.1 & 23.0 & 24.2 & 35.6 & 7.70 & 36.8 \\
\midrule
LLaVA1.5-7B-SFT & 35.9 & 28.4 & 21.0 & 25.0 & 33.1 & 7.70 & 33.3\\
LLaVA1.5-13B-SFT  & 44.3 & 27.0 & 28.1 & 29.9 & 32.1 & 	7.70 & 38.3\\
\midrule
\textbf{LLaVA1.5-7B-BPO}  & \textbf{41.3} & \textbf{29.5} & \textbf{24.8} & \textbf{27.8} & \textbf{34.8} & \textbf{11.5} & \textbf{36.8}\\
\textbf{LLaVA1.5-13B-BPO} & \textbf{46.9} & \textbf{31.6} & \textbf{34.6} & \textbf{37.2} & \textbf{36.1} & \textbf{11.5} &\textbf{41.4}\\
\bottomrule
\end{tabular}
}
\label{tab:sft}
\vspace{-0.35cm}
\end{table*}
\subsection{Comparison with SFT}
A straightforward baseline approach would involve supervised fine-tuning, aiming to address the questions: ``How effective is the preference learning algorithm?'' and ``How significant are the negative responses?'' To validate this, we extract solely the positive responses from our preference dataset and proceed with SFT. As demonstrated by the results in table~\ref{tab:sft}, we notice only a marginal improvement in performance compared to the baseline methods. This demonstrates the indispensability of negative responses and preference learning.
\subsection{Comparison with Self-generated Response}
Another straightforward question is whether signifying the pre-training bias in MLLMs is necessary. We compare the results achieved on LLaVA-7B by directly using the responses bootstrapped from MLLMs as negative samples with responses generated using image-weakened prompting in table~\ref{tab:self-gen}, which verifies that purposefully exposing pretraining bias helps achieve better performance.
\begin{table*}[htp!]
    \centering
\caption{Comparison with self-generated responses without image weakening. 
}
{
\begin{tabular}{l!{\vrule width 0.5pt}cccccccc}

\toprule
Model & Rec & OCR & Know & Gen & Spat & Math & Total\\
\midrule
Baseline & 37.0 & 22.9 &16.8  & 20.2 &25.7 &7.70 &31.7 \\
\midrule
Self-generated & 38.6 &	21.7 &	21.5 &	20.8 &	28.3 &	1.90 &	32.4\\
\midrule
\textbf{Image-Weakened}  & \textbf{38.9} & \textbf{23.5} & \textbf{24.8} & \textbf{21.2} & \textbf{28.5} & \textbf{7.70} & \textbf{34.3} \\
\bottomrule
\end{tabular}
}
\label{tab:self-gen}
\end{table*}

\begin{figure*}[htp!]
\centering
\includegraphics[width=1\textwidth]{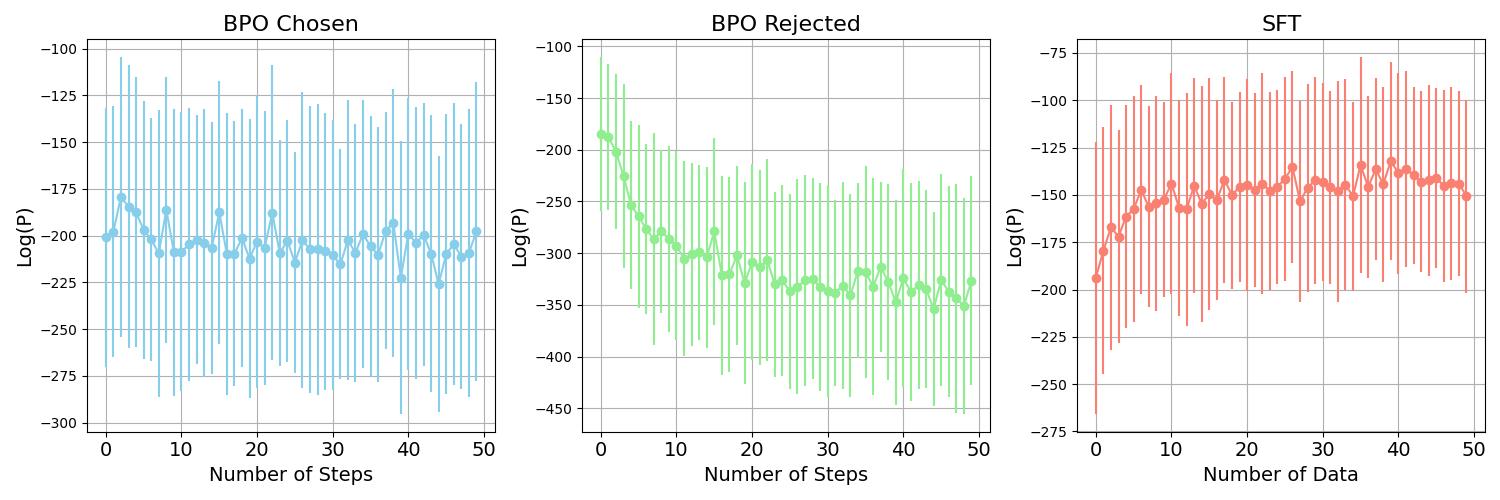} 
\vspace{-0.8cm}
\caption{The value of log probabilities for responses throughout the training process. Left: positive responses for \ourmodel; Middle: negative responses  for \ourmodel ; Right: ground truth responses for supervised fine-tuning (SFT).   }\label{fig:log_prob}
\vspace{-0.7cm}
\end{figure*}
\subsection{Comparison of Loss Curve}
Figure~\ref{fig:log_prob} visualizes the log probabilities of responses throughout the training process. The left graph displays the log probabilities corresponding to positive responses during the training of our model. In the middle graph, the log probabilities for negative responses during the same training phase are presented. On the right side, the log probabilities for ground truth responses during supervised fine-tuning (SFT) are shown. We plot the mean value of every interval spanning 100 steps, and add visualize the standard deviation with the error bars. 

It is noteworthy that the log probability for negative responses consistently decreases during the process of \ourmodel, while the log probability for positive responses remains relatively stable. Conversely, the log probability for ground truth responses steadily increases for SFT. This observation suggests that preference learning aims to establish a clear margin between the likelihoods of positive and negative responses, effectively suppressing the biases originating from pretraining. On the other hand, the supervised fine-tuning (SFT) approach aims to memorize the ground truth annotations, which continuously increases the likelihood of ground truth responses, potentially leading to overfitting and the occurrence of catastrophic forgetting~\cite{lin2023speciality}.
\begin{figure*}[htp!]
\centering
\includegraphics[width=0.7\textwidth]{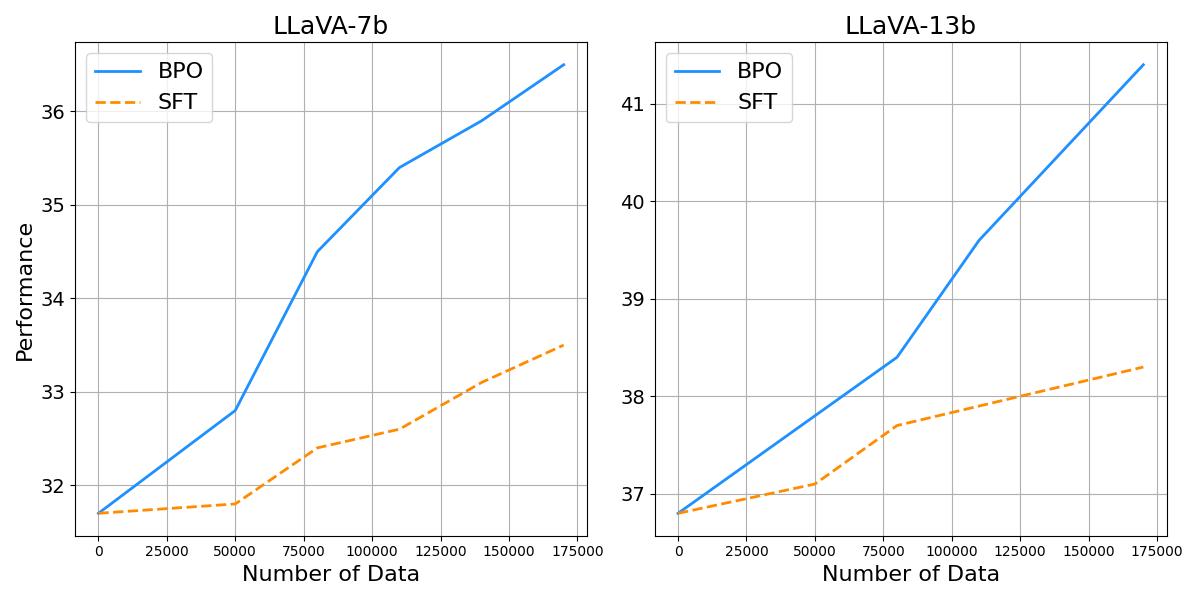} 
\vspace{-0.3cm}
\caption{We show the performance gain introduced by BPO and SFT on LLaVA-7B and LLaVA-13B, respectively. BPO  consistently outperforms SFT across all dataset sizes.}\label{fig:size}
\end{figure*}
\subsection{Effect of Dataset Sizes}
We examine the improvement brought by \ourmodel with various sizes for preference dataset and compared to that of supervised fine-tuning in table~\ref{fig:size}. We find that larger scale of preference data indeed leads to better performance. In addition, the effectiveness of \ourmodel consistently dominate that of SFT, which verifies that preference learning has better sample efficiency than the SFT counterpart.
%
\label{sec:experiments}
\section{Conclusion}
In conclusion, our paper introduces Bootstrapped Preference Optimization (\ourmodel) as a solution to mitigate bias in Multimodal Large Language Models (LLMs) when generating responses based on visual inputs. By curating paired preference datasets through bootstrapping negative responses from the model itself, we encourage the model's grounding in visual information by suppressing the preference for pretraining bias. Our approach leads to significant performance improvements across multiple benchmarks and advancing the state-of-the-art in multimodal conversational systems. We hope that our method will encourage more future research towards stronger multimodal alignment.
\label{sec:conclusion}
\section{Limitations}
In this paper, we verified that our bootstrapped preference optmization (BPO) techinque is able  to effectively improve the model's performance. However, we still have the following unresolved issues: 1) We do not take into considerations of the malicious inputs that may cause  to  model to produce harmful responses~\cite{liu2023prompt, kasirzadeh2022conversation, bai2022training, xie2023defending, shen2023anything, liu2023jailbreaking, yi2023benchmarking, xie2024gradsafe, qi2023fine,yi2024opensource}, which may lead to undesired consequences; 2) we do not design sample selection algorithms for filtering out data potentially with low quality~\cite{yu2019does, han2018coteaching, xia2019anchor, gao2023selfguided, pmlr-v162-zhou22h, Lin2023AHV}. We will leave those unresolved  issues to future research.
\label{sec:limitations}
\par\vfill\par

%
%
\bibliographystyle{splncs04}
\bibliography{main}

\begin{thebibliography}{10}
\providecommand{\url}[1]{\texttt{#1}}
\providecommand{\urlprefix}{URL }
\providecommand{\doi}[1]{https://doi.org/#1}

\bibitem{bai2023qwenvl}
Bai, J., Bai, S., Yang, S., Wang, S., Tan, S., Wang, P., Lin, J., Zhou, C., Zhou, J.: Qwen-vl: A versatile vision-language model for understanding, localization, text reading, and beyond (2023)

\bibitem{bai2022training}
Bai, Y., Jones, A., Ndousse, K., Askell, A., Chen, A., DasSarma, N., Drain, D., Fort, S., Ganguli, D., Henighan, T., et~al.: Training a helpful and harmless assistant with reinforcement learning from human feedback. arXiv preprint arXiv:2204.05862  (2022)

\bibitem{bai2022constitutional}
Bai, Y., Kadavath, S., Kundu, S., Askell, A., Kernion, J., Jones, A., Chen, A., Goldie, A., Mirhoseini, A., McKinnon, C., et~al.: Constitutional ai: Harmlessness from ai feedback. arXiv preprint arXiv:2212.08073  (2022)

\bibitem{bradley1952rank}
Bradley, R.A., Terry, M.E.: Rank analysis of incomplete block designs: I. the method of paired comparisons. Biometrika  \textbf{39}(3/4),  324--345 (1952)

\bibitem{brown2020language}
Brown, T., Mann, B., Ryder, N., Subbiah, M., Kaplan, J.D., Dhariwal, P., Neelakantan, A., Shyam, P., Sastry, G., Askell, A., et~al.: Language models are few-shot learners. Advances in neural information processing systems  \textbf{33},  1877--1901 (2020)

\bibitem{chen2023sharegpt4v}
Chen, L., Li, J., Dong, X., Zhang, P., He, C., Wang, J., Zhao, F., Lin, D.: Sharegpt4v: Improving large multi-modal models with better captions (2023)

\bibitem{chen2020simple}
Chen, T., Kornblith, S., Norouzi, M., Hinton, G.: A simple framework for contrastive learning of visual representations (2020)

\bibitem{vicuna2023}
Chiang, W.L., Li, Z., Lin, Z., Sheng, Y., Wu, Z., Zhang, H., Zheng, L., Zhuang, S., Zhuang, Y., Gonzalez, J.E., Stoica, I., Xing, E.P.: Vicuna: An open-source chatbot impressing gpt-4 with 90\%* chatgpt quality (March 2023), \url{https://lmsys.org/blog/2023-03-30-vicuna/}

\bibitem{choshen2019weaknesses}
Choshen, L., Fox, L., Aizenbud, Z., Abend, O.: On the weaknesses of reinforcement learning for neural machine translation. arXiv preprint arXiv:1907.01752  (2019)

\bibitem{chowdhery2022palm}
Chowdhery, A., Narang, S., Devlin, J., Bosma, M., Mishra, G., Roberts, A., Barham, P., Chung, H.W., Sutton, C., Gehrmann, S., et~al.: Palm: Scaling language modeling with pathways. arXiv preprint arXiv:2204.02311  (2022)

\bibitem{christiano2017deep}
Christiano, P.F., Leike, J., Brown, T., Martic, M., Legg, S., Amodei, D.: Deep reinforcement learning from human preferences. Advances in neural information processing systems  \textbf{30} (2017)

\bibitem{dai2023instructblip}
Dai, W., Li, J., Li, D., Tiong, A.M.H., Zhao, J., Wang, W., Li, B., Fung, P., Hoi, S.: Instructblip: Towards general-purpose vision-language models with instruction tuning (2023)

\bibitem{diao2023lmflow}
Diao, S., Pan, R., Dong, H., Shum, K.S., Zhang, J., Xiong, W., Zhang, T.: Lmflow: An extensible toolkit for finetuning and inference of large foundation models. arXiv preprint arXiv:2306.12420  (2023)

\bibitem{ding2023hilm}
Ding, X., Han, J., Xu, H., Zhang, W., Li, X.: Hilm-d: Towards high-resolution understanding in multimodal large language models for autonomous driving. arXiv preprint arXiv:2309.05186  (2023)

\bibitem{ding2023hilmd}
Ding, X., Han, J., Xu, H., Zhang, W., Li, X.: Hilm-d: Towards high-resolution understanding in multimodal large language models for autonomous driving (2023)

\bibitem{dong2023raft}
Dong, H., Xiong, W., Goyal, D., Zhang, Y., Chow, W., Pan, R., Diao, S., Zhang, J., Shum, K., Zhang, T.: Raft: Reward ranked finetuning for generative foundation model alignment (2023)

\bibitem{douglas2024mitigating}
Douglas, R., Draguns, A., Gavenčiak, T.: Mitigating the problem of strong priors in lms with context extrapolation (2024)

\bibitem{gao2023selfguided}
Gao, J., Pi, R., Lin, Y., Xu, H., Ye, J., Wu, Z., Zhang, W., Liang, X., Li, Z., Kong, L.: Self-guided noise-free data generation for efficient zero-shot learning (2023)

\bibitem{gao2023gllava}
Gao, J., Pi, R., Zhang, J., Ye, J., Zhong, W., Wang, Y., Hong, L., Han, J., Xu, H., Li, Z., Kong, L.: G-llava: Solving geometric problem with multi-modal large language model (2023)

\bibitem{gao2023llamaadapter}
Gao, P., Han, J., Zhang, R., Lin, Z., Geng, S., Zhou, A., Zhang, W., Lu, P., He, C., Yue, X., Li, H., Qiao, Y.: Llama-adapter v2: Parameter-efficient visual instruction model (2023)

\bibitem{openlm2023openllama}
Geng, X., Liu, H.: Openllama: An open reproduction of llama (May 2023), \url{https://github.com/openlm-research/open_llama}

\bibitem{glaese2022improving}
Glaese, A., McAleese, N., Tr{\k{e}}bacz, M., Aslanides, J., Firoiu, V., Ewalds, T., Rauh, M., Weidinger, L., Chadwick, M., Thacker, P., et~al.: Improving alignment of dialogue agents via targeted human judgements. arXiv preprint arXiv:2209.14375  (2022)

\bibitem{gemini2023}
Google: Gemini: A family of highly capable multimodal models (2023), \url{https://storage.googleapis.com/deepmind-media/gemini/gemini_1_report.pdf}

\bibitem{han2018coteaching}
Han, B., Yao, Q., Yu, X., Niu, G., Xu, M., Hu, W., Tsang, I., Sugiyama, M.: Co-teaching: Robust training of deep neural networks with extremely noisy labels (2018)

\bibitem{han2024instinctive}
Han, T., Lian, Q., Pan, R., Pi, R., Zhang, J., Diao, S., Lin, Y., Zhang, T.: The instinctive bias: Spurious images lead to hallucination in mllms (2024)

\bibitem{hao2022optimizing}
Hao, Y., Chi, Z., Dong, L., Wei, F.: Optimizing prompts for text-to-image generation. arXiv preprint arXiv:2212.09611  (2022)

\bibitem{ho2020denoising}
Ho, J., Jain, A., Abbeel, P.: Denoising diffusion probabilistic models (2020)

\bibitem{hoffmann2022training}
Hoffmann, J., Borgeaud, S., Mensch, A., Buchatskaya, E., Cai, T., Rutherford, E., Casas, D.d.L., Hendricks, L.A., Welbl, J., Clark, A., et~al.: Training compute-optimal large language models. arXiv preprint arXiv:2203.15556  (2022)

\bibitem{kasirzadeh2022conversation}
Kasirzadeh, A., Gabriel, I.: In conversation with artificial intelligence: aligning language models with human values. arXiv preprint arXiv:2209.00731  (2022)

\bibitem{lee2023aligning}
Lee, K., Liu, H., Ryu, M., Watkins, O., Du, Y., Boutilier, C., Abbeel, P., Ghavamzadeh, M., Gu, S.S.: Aligning text-to-image models using human feedback. arXiv preprint arXiv:2302.12192  (2023)

\bibitem{leike2018scalable}
Leike, J., Krueger, D., Everitt, T., Martic, M., Maini, V., Legg, S.: Scalable agent alignment via reward modeling: a research direction. arXiv preprint arXiv:1811.07871  (2018)

\bibitem{leng2023mitigating}
Leng, S., Zhang, H., Chen, G., Li, X., Lu, S., Miao, C., Bing, L.: Mitigating object hallucinations in large vision-language models through visual contrastive decoding (2023)

\bibitem{li2023otter}
Li, B., Zhang, Y., Chen, L., Wang, J., Yang, J., Liu, Z.: Otter: A multi-modal model with in-context instruction tuning (2023)

\bibitem{li2023llavamed}
Li, C., Wong, C., Zhang, S., Usuyama, N., Liu, H., Yang, J., Naumann, T., Poon, H., Gao, J.: Llava-med: Training a large language-and-vision assistant for biomedicine in one day (2023)

\bibitem{li2023blip2}
Li, J., Li, D., Savarese, S., Hoi, S.: Blip-2: Bootstrapping language-image pre-training with frozen image encoders and large language models (2023)

\bibitem{li2023silkie}
Li, L., Xie, Z., Li, M., Chen, S., Wang, P., Chen, L., Yang, Y., Wang, B., Kong, L.: Silkie: Preference distillation for large visual language models (2023)

\bibitem{li2023evaluating}
Li, Y., Du, Y., Zhou, K., Wang, J., Zhao, W.X., Wen, J.R.: Evaluating object hallucination in large vision-language models (2023)

\bibitem{Lin2023AHV}
Lin, Y., Pi, R., Zhang, W., Xia, X., Gao, J., Zhou, X., Liu, T., Han, B.: A holistic view of label noise transition matrix in deep learning and beyond. In: International Conference on Learning Representations (2023), \url{https://api.semanticscholar.org/CorpusID:259298577}

\bibitem{lin2023speciality}
Lin, Y., Tan, L., Lin, H., Zheng, Z., Pi, R., Zhang, J., Diao, S., Wang, H., Zhao, H., Yao, Y., et~al.: Speciality vs generality: An empirical study on catastrophic forgetting in fine-tuning foundation models. arXiv preprint arXiv:2309.06256  (2023)

\bibitem{liu2023mitigating}
Liu, F., Lin, K., Li, L., Wang, J., Yacoob, Y., Wang, L.: Mitigating hallucination in large multi-modal models via robust instruction tuning (2023)

\bibitem{liu2023llava}
Liu, H., Li, C., Wu, Q., Lee, Y.J.: Visual instruction tuning (2023)

\bibitem{liu2023queryrelevant}
Liu, X., Zhu, Y., Lan, Y., Yang, C., Qiao, Y.: Query-relevant images jailbreak large multi-modal models (2023)

\bibitem{liu2023prompt}
Liu, Y., Deng, G., Li, Y., Wang, K., Zhang, T., Liu, Y., Wang, H., Zheng, Y., Liu, Y.: Prompt injection attack against llm-integrated applications. arXiv preprint arXiv:2306.05499  (2023)

\bibitem{liu2023jailbreaking}
Liu, Y., Deng, G., Xu, Z., Li, Y., Zheng, Y., Zhang, Y., Zhao, L., Zhang, T., Liu, Y.: Jailbreaking chatgpt via prompt engineering: An empirical study. arXiv preprint arXiv:2305.13860  (2023)

\bibitem{meng2022generating}
Meng, Y., Huang, J., Zhang, Y., Han, J.: Generating training data with language models: Towards zero-shot language understanding. arXiv preprint arXiv:2202.04538  (2022)

\bibitem{meng2023tuning}
Meng, Y., Michalski, M., Huang, J., Zhang, Y., Abdelzaher, T., Han, J.: Tuning language models as training data generators for augmentation-enhanced few-shot learning (2023)

\bibitem{nakano2021webgpt}
Nakano, R., Hilton, J., Balaji, S., Wu, J., Ouyang, L., Kim, C., Hesse, C., Jain, S., Kosaraju, V., Saunders, W., et~al.: Webgpt: Browser-assisted question-answering with human feedback. arXiv preprint arXiv:2112.09332  (2021)

\bibitem{openai2023gpt4}
OpenAI: Gpt-4 technical report (2023)

\bibitem{ouyang2022training}
Ouyang, L., Wu, J., Jiang, X., Almeida, D., Wainwright, C., Mishkin, P., Zhang, C., Agarwal, S., Slama, K., Ray, A., et~al.: Training language models to follow instructions with human feedback. Advances in Neural Information Processing Systems  \textbf{35},  27730--27744 (2022)

\bibitem{pi2023detgpt}
Pi, R., Gao, J., Diao, S., Pan, R., Dong, H., Zhang, J., Yao, L., Han, J., Xu, H., Kong, L., Zhang, T.: Detgpt: Detect what you need via reasoning (2023)

\bibitem{pi2024mllmprotector}
Pi, R., Han, T., Xie, Y., Pan, R., Lian, Q., Dong, H., Zhang, J., Zhang, T.: Mllm-protector: Ensuring mllm's safety without hurting performance (2024)

\bibitem{pi2023perceptiongpt}
Pi, R., Yao, L., Gao, J., Zhang, J., Zhang, T.: Perceptiongpt: Effectively fusing visual perception into llm (2023)

\bibitem{qi2023fine}
Qi, X., Zeng, Y., Xie, T., Chen, P.Y., Jia, R., Mittal, P., Henderson, P.: Fine-tuning aligned language models compromises safety, even when users do not intend to! arXiv preprint arXiv:2310.03693  (2023)

\bibitem{rafailov2023direct}
Rafailov, R., Sharma, A., Mitchell, E., Ermon, S., Manning, C.D., Finn, C.: Direct preference optimization: Your language model is secretly a reward model (2023)

\bibitem{ramamurthy2022reinforcement}
Ramamurthy, R., Ammanabrolu, P., Brantley, K., Hessel, J., Sifa, R., Bauckhage, C., Hajishirzi, H., Choi, Y.: Is reinforcement learning (not) for natural language processing?: Benchmarks, baselines, and building blocks for natural language policy optimization. arXiv preprint arXiv:2210.01241  (2022)

\bibitem{rohrbach2019object}
Rohrbach, A., Hendricks, L.A., Burns, K., Darrell, T., Saenko, K.: Object hallucination in image captioning (2019)

\bibitem{scao2022bloom}
Scao, T.L., Fan, A., Akiki, C., Pavlick, E., Ili{c}, S., Hesslow, D., Castagn{e}, R., Luccioni, A.S., Yvon, F., Gall{e}, M., et~al.: Bloom: A 176b-parameter open-access multilingual language model. arXiv preprint arXiv:2211.05100  (2022)

\bibitem{scheurer2023training}
Scheurer, J., Campos, J.A., Korbak, T., Chan, J.S., Chen, A., Cho, K., Perez, E.: Training language models with language feedback at scale. arXiv preprint arXiv:2303.16755  (2023)

\bibitem{schulman2017proximal}
Schulman, J., Wolski, F., Dhariwal, P., Radford, A., Klimov, O.: Proximal policy optimization algorithms (2017)

\bibitem{shen2023anything}
Shen, X., Chen, Z., Backes, M., Shen, Y., Zhang, Y.: " do anything now": Characterizing and evaluating in-the-wild jailbreak prompts on large language models. arXiv preprint arXiv:2308.03825  (2023)

\bibitem{smith2022using}
Smith, S., Patwary, M., Norick, B., LeGresley, P., Rajbhandari, S., Casper, J., Liu, Z., Prabhumoye, S., Zerveas, G., Korthikanti, V., et~al.: Using deepspeed and megatron to train megatron-turing nlg 530b, a large-scale generative language model. arXiv preprint arXiv:2201.11990  (2022)

\bibitem{stiennon2020learning}
Stiennon, N., Ouyang, L., Wu, J., Ziegler, D., Lowe, R., Voss, C., Radford, A., Amodei, D., Christiano, P.F.: Learning to summarize with human feedback. Advances in Neural Information Processing Systems  \textbf{33},  3008--3021 (2020)

\bibitem{su2023pandagpt}
Su, Y., Lan, T., Li, H., Xu, J., Wang, Y., Cai, D.: Pandagpt: One model to instruction-follow them all (2023)

\bibitem{sun2023aligning}
Sun, Z., Shen, S., Cao, S., Liu, H., Li, C., Shen, Y., Gan, C., Gui, L.Y., Wang, Y.X., Yang, Y., Keutzer, K., Darrell, T.: Aligning large multimodal models with factually augmented rlhf (2023)

\bibitem{alpaca}
Taori, R., Gulrajani, I., Zhang, T., Dubois, Y., Li, X., Guestrin, C., Liang, P., Hashimoto, T.B.: Stanford alpaca: An instruction-following llama model. \url{https://github.com/tatsu-lab/stanford_alpaca} (2023)

\bibitem{touvron2023llama}
Touvron, H., Lavril, T., Izacard, G., Martinet, X., Lachaux, M.A., Lacroix, T., Rozi{\`e}re, B., Goyal, N., Hambro, E., Azhar, F., et~al.: Llama: Open and efficient foundation language models. arXiv preprint arXiv:2302.13971  (2023)

\bibitem{wang2024arithmetic}
Wang, H., Lin, Y., Xiong, W., Yang, R., Diao, S., Qiu, S., Zhao, H., Zhang, T.: Arithmetic control of llms for diverse user preferences: Directional preference alignment with multi-objective rewards. arXiv preprint arXiv:2402.18571  (2024)

\bibitem{wu2021recursively}
Wu, J., Ouyang, L., Ziegler, D.M., Stiennon, N., Lowe, R., Leike, J., Christiano, P.: Recursively summarizing books with human feedback. arXiv preprint arXiv:2109.10862  (2021)

\bibitem{wu2023better}
Wu, X., Sun, K., Zhu, F., Zhao, R., Li, H.: Better aligning text-to-image models with human preference. arXiv preprint arXiv:2303.14420  (2023)

\bibitem{xia2019anchor}
Xia, X., Liu, T., Wang, N., Han, B., Gong, C., Niu, G., Sugiyama, M.: Are anchor points really indispensable in label-noise learning? (2019)

\bibitem{xie2024gradsafe}
Xie, Y., Fang, M., Pi, R., Gong, N.: Gradsafe: Detecting unsafe prompts for llms via safety-critical gradient analysis (2024)

\bibitem{xie2023defending}
Xie, Y., Yi, J., Shao, J., Curl, J., Lyu, L., Chen, Q., Xie, X., Wu, F.: Defending chatgpt against jailbreak attack via self-reminders. Nature Machine Intelligence pp. 1--11 (2023)

\bibitem{xiong2023gibbs}
Xiong, W., Dong, H., Ye, C., Zhong, H., Jiang, N., Zhang, T.: Gibbs sampling from human feedback: A provable kl-constrained framework for rlhf. arXiv preprint arXiv:2312.11456  (2023)

\bibitem{ye2022zerogen}
Ye, J., Gao, J., Li, Q., Xu, H., Feng, J., Wu, Z., Yu, T., Kong, L.: Zerogen: Efficient zero-shot learning via dataset generation. In: Empirical Methods in Natural Language Processing (2022)

\bibitem{yi2023benchmarking}
Yi, J., Xie, Y., Zhu, B., Hines, K., Kiciman, E., Sun, G., Xie, X., Wu, F.: Benchmarking and defending against indirect prompt injection attacks on large language models. arXiv preprint arXiv:2312.14197  (2023)

\bibitem{yi2024opensource}
Yi, J., Ye, R., Chen, Q., Zhu, B.B., Chen, S., Lian, D., Sun, G., Xie, X., Wu, F.: Open-source can be dangerous: On the vulnerability of value alignment in open-source {LLM}s. \url{https://openreview.net/forum?id=NIouO0C0ex} (2024)

\bibitem{yin2023woodpecker}
Yin, S., Fu, C., Zhao, S., Xu, T., Wang, H., Sui, D., Shen, Y., Li, K., Sun, X., Chen, E.: Woodpecker: Hallucination correction for multimodal large language models (2023)

\bibitem{yu2023metamath}
Yu, L., Jiang, W., Shi, H., Yu, J., Liu, Z., Zhang, Y., Kwok, J.T., Li, Z., Weller, A., Liu, W.: Metamath: Bootstrap your own mathematical questions for large language models (2023)

\bibitem{yu2023rlhfv}
Yu, T., Yao, Y., Zhang, H., He, T., Han, Y., Cui, G., Hu, J., Liu, Z., Zheng, H.T., Sun, M., Chua, T.S.: Rlhf-v: Towards trustworthy mllms via behavior alignment from fine-grained correctional human feedback (2023)

\bibitem{yu2023mmvet}
Yu, W., Yang, Z., Li, L., Wang, J., Lin, K., Liu, Z., Wang, X., Wang, L.: Mm-vet: Evaluating large multimodal models for integrated capabilities (2023)

\bibitem{yu2019does}
Yu, X., Han, B., Yao, J., Niu, G., Tsang, I.W., Sugiyama, M.: How does disagreement help generalization against label corruption? (2019)

\bibitem{yuan2023rrhf}
Yuan, Z., Yuan, H., Tan, C., Wang, W., Huang, S., Huang, F.: Rrhf: Rank responses to align language models with human feedback without tears. arXiv preprint arXiv:2304.05302  (2023)

\bibitem{zhang_2023_ltbook}
Zhang, T.: Mathematical Analysis of Machine Learning Algorithms. Cambridge University Press (2023). \doi{10.1017/9781009093057}

\bibitem{zhang2024llavar}
Zhang, Y., Zhang, R., Gu, J., Zhou, Y., Lipka, N., Yang, D., Sun, T.: Llavar: Enhanced visual instruction tuning for text-rich image understanding (2024)

\bibitem{pmlr-v162-zhou22h}
Zhou, X., Pi, R., Zhang, W., Lin, Y., Chen, Z., Zhang, T.: Probabilistic bilevel coreset selection. In: Chaudhuri, K., Jegelka, S., Song, L., Szepesvari, C., Niu, G., Sabato, S. (eds.) Proceedings of the 39th International Conference on Machine Learning. Proceedings of Machine Learning Research, vol.~162, pp. 27287--27302. PMLR (17--23 Jul 2022), \url{https://proceedings.mlr.press/v162/zhou22h.html}

\bibitem{zhu2023minigpt4}
Zhu, D., Chen, J., Shen, X., Li, X., Elhoseiny, M.: Minigpt-4: Enhancing vision-language understanding with advanced large language models (2023)

\bibitem{ziegler2019fine}
Ziegler, D.M., Stiennon, N., Wu, J., Brown, T.B., Radford, A., Amodei, D., Christiano, P., Irving, G.: Fine-tuning language models from human preferences. arXiv preprint arXiv:1909.08593  (2019)

\end{thebibliography}
\end{document}